\documentclass{svproc}

\usepackage{url}

\usepackage{graphicx}
\usepackage{multicol}
\usepackage{footmisc}
\usepackage{color}
\usepackage{hyperref}
\usepackage{amsmath}

\begin{document}
\mainmatter              
\title{Learning Nonlinearity of Boolean Functions\\ 
-- An Experimentation with Neural Networks}
\titlerunning{Learning Nonlinearity of Boolean Functions}

\author{Sriram Ranga \and
Nandish Chattopadhyay\and
Anupam Chattopadhyay}

\authorrunning{S. Ranga et al.}
\tocauthor{Sriram Ranga, Nandish Chattopadhyay and Anupam Chattopadhyay}
\institute{Nanyang Technological University, 50 Nanyang Ave, Singapore 639798\\
\email{sriram011@e.ntu.edu.sg}}

\maketitle              
\begin{abstract}
This paper investigates the learnability of the nonlinearity property of Boolean functions using neural networks. We train encoder style deep neural networks to learn to predict the nonlinearity of Boolean functions from examples of functions in the form of a truth table and their corresponding nonlinearity values. We report empirical results to show that deep neural networks are able to learn to predict the property for functions in 4 and 5 variables with an accuracy above 95\%. While these results are positive and a disciplined analysis is being presented for the first time in this regard, we should also underline the statutory warning that it seems quite challenging to extend the idea to higher number of variables, and it is also not clear whether one can get advantage in terms of time and space complexity over the existing combinatorial algorithms.
\end{abstract}

\section{Introduction}

Boolean functions are the building blocks of cryptographic systems \cite{wu_boolean_2016}, and these combinatorial objects have important applications in many other domains of science and technology. They are used in designing different kinds of ciphers and it is imperative that they have the required properties so that the designer can be confident that the systems are immune to cryptanalytic attacks. Nonlinearity, correlation immunity, balancedness, and resiliency are among the most important of such properties \cite{SarkarMaitraConstruction}. In this paper, we contribute to this research domain by studying the learnability of the nonlinearity property from examples, using neural networks.

Traditional, deterministic algorithms exist to calculate all the properties mentioned above, but non-classical algorithms have shown that there is scope for improvement \cite{deutsch_jozsa_algorithm}\cite{PhysRevA.84.062329}. In fact, to determine whether a given Boolean function is balanced or constant, is one of the first examples of quantum algorithms that is exponentially faster than a deterministic classical algorithm. The fields of cryptography and machine learning have  contributed multiple ideas and techniques to each other \cite{goos_cryptography_1993}. While neural networks, if run on a classical computer, should be considered as classical heuristics, they were not extensively studied in estimating nonlinearity of Boolean functions. Research on the use of neural networks in Boolean function operations like the inner product \cite{erdal_learnability_2022} and properties like parity \cite{HOHIL19991321}\cite{Furst1981ParityCA} is generally focused on the computational complexity \cite{ian_circuit_complexity}\cite{nn_boolean_model} of simple hand crafted networks designed using threshold neurons. In contrast to that, the use of deep neural networks has recently shown strong positive results in the learnability of Boolean functions themselves \cite{learnability_dnn_vardi}. The discrete nature of Boolean functions and that of the mathematical operations required to calculate nonlinearity make a strong case for the position that a gradient descent based solution for the problem might not be possible. Traditionally, machine learning approaches are preferred over regular algorithms in tasks like textual sentiment analysis and image recognition which are probabilistic in nature and possibly ridden with ambiguity.
The absence of such uncertainty in this deterministic problem of learning Boolean function properties presents us with an opportunity to explore the possibility of a fully interpretable and perfectly generalizing neural network . This problem is thus in the intersection of Boolean functions, and learnability and interpretability of neural networks. This paper discusses it from all these various perspectives.

\subsection{Contribution and Organization}
We first show that the sub-problem of calculating the Walsh spectrum of a Boolean function can be learnt by a neural network with a single hidden layer (see Fig. \ref{fig1}). This simple linear network, when given examples of Boolean functions and corresponding Walsh spectra, can learn to perform the correct affine transformation on any $n$ variable Boolean function, for which it is required to calculate its Walsh spectrum. 
We observe that this can be achieved using a minimum of $2^n$ functions and their corresponding Walsh spectra as examples. The parameters of the network after training resemble the Walsh Hadamard matrix, which gives us visual proof that the network has learn to solve the problem well. We like to remind here that construction of Walsh Hadamard matrix in this regard can also be achieved by solving a set of equations given the same number, i.e., $2^n$ functions in the form of a truth table.  

We then move on to the specific problem of learning nonlinearity in an end-to-end fashion. We find that this problem can be modeled using a network similar to the one used earlier, but it does not converge to the correct set of weights when trained using examples. When wider and deeper encoder style neural networks were used instead, the networks after training could predict nonlinearity with more than 95\% accuracy for 4 and 5 variable Boolean functions. However, we fail to extend these results to functions in higher number of variables. Additionally, we show that these networks are less memory and time efficient compared to the traditional algorithms used for calculating nonlinearity. In spite of this, it is encouraging that on small number of variables, the results are quite close and certain further ideas need to be explored in this regard.

The paper is organized as follows. \hyperlink{section2}{Section 2} covers the basics of Boolean functions and the different methods to calculate their nonlinearity. \hyperlink{section3}{Section 3} starts off with a comparison of the common methods for calculating nonlinearity and sets the expectations for what could be considered a good neural network based solution for predicting nonlinearity. \hyperlink{section3_1}{Section 3.1} discusses the learnability of the Walsh spectrum and \hyperlink{section3_2}{Section 3.2} has the same for end to end learnability of nonlinearity. We conclude in \hyperlink{section4}{Section 4} with a short note on potential future work. The code used for the experiments is made available \hypertarget{section2}{in} \href{https://github.com/sriram-ranga-ntu/nn-boolean}{GitHub}.

\section{Technical Background}
In this section, we explain different combinatorial aspects of Boolean functions that are generally used in the domain of cryptology.

\subsection{Boolean Functions}

\begin{definition}[Boolean Function]
    An $n$-variable Boolean function is a function $f: X^n \rightarrow X$ where $X = \{0, 1\} $.
\end{definition}

Let us denote $2^n$ by $N$. An $n$ variable Boolean function can take $N$ inputs. The function can be thus represented as a string of $N$ 1s or 0s depending on the truth value of the function for each of the $N$ possible inputs ordered lexicographically.

\newtheorem{prop}{Proposition}
\begin{prop}
    There exist $2^{N}$ $n$-variable Boolean functions.
\end{prop}

The domain of an $n$-variable Boolean function has $N$ elements, and the co-domain is the 2-element set \{0,1\}. Thus there exist a total of $2^N$ (or $2^{2^n}$) $n$-variable Boolean functions.

The function space of Boolean functions increases super-exponentially with the number of variables, and this gives the advantage of making it very difficult for an attacker to guess the exact Boolean function which was used in any cryptography system.

\begin{definition}[Weight of a Boolean Function]
    The weight of a Boolean function $f: \{0, 1\}^n \rightarrow \{0, 1\}$, denoted by $wt(f)$ is a scalar value equal to the number of inputs for which the output equals 1.
\end{definition}

\begin{definition}[Hamming Distance]
    Hamming distance between two Boolean functions $f_1$ and $f_2$, denoted as $d(f_1, f_2)$, is defined as $wt(f_1 \oplus f_2)$.
\end{definition}
 Hamming Distance between two functions can be thought of as the number of places in their truth tables in which they differ from each other.

\begin{definition}[Algebraic Normal Form]
    An $n$-variable Boolean function $f(x_1, x_2, ... x_n)$ can be written as 
    \begin{equation}
        a_0 \oplus \underset{1 \leq i \leq n}{\bigoplus} a_ix_i \oplus \underset{1 \leq i<j \leq n}{\bigoplus} a_{ij}x_ix_j \oplus \bigoplus \dots
 \oplus a_{12\dots n}x_1x_2\dots x_n
    \end{equation}
    where each coefficient can take a value of 0 or 1. A function written in this form, as a sum of all distinct $k^{th}$ order products, is said to be in its Algebraic Normal Form (ANF).
 
\end{definition}

\begin{definition}[Algebraic Degree of a Boolean Function]

For a Boolean function $f$, expressed in its ANF, the algebraic degree or simply degree of $f$, denoted by $deg(f)$ is defined as the number of variables in the highest product term of $f$ with a non-zero coefficient. 
    
\end{definition}
Note: The degree of a zero function is considered to be 0.

\begin{definition}[Affine Function]

Boolean functions of degree at most one are called affine functions.
\end{definition}

An affine function takes the form \begin{equation}
    a_0 \oplus a_1 x_1 \oplus ... \oplus a_n x_n    
\end{equation}
where each $a_i$ can take one value from \{0,1\}. There are thus $2N$ (or $2^{n+1}$) $n$-variable affine functions. The set of all $n$-variable affine functions is denoted by the set $A(n)$. Observe that if the value of the constant term $a_0$ is changed keeping all other constants and variables the same, the output of the function flips since $(!a)\oplus b = !(a\oplus b)$. Thus we can think of $A(n)$ as $N$ pairs of affine functions with exactly opposite truth tables.

\subsection{Nonlinearity}

\begin{definition}[Nonlinearity of a Boolean Function]
    The nonlinearity of an $n$-variable Boolean function $f$, denoted by $nl(f)$ is defined as the minimum hamming distance of $f$ from the set of all $n$ variable affine functions.
    
    \begin{equation}
        nl(f) = \underset{g \in A(n)}{min}(d(f, g))
    \end{equation}
    
\end{definition}

Functions with higher nonlinearity are preferred in cryptography since it is difficult to approximate highly nonlinear functions with simple affine functions. 

\begin{definition}[Walsh Transform]
    Let $x = (x_1, ... , x_n)$ and $\omega = (\omega_1, ..., \omega_n)$ both belong to $\{0,1\}^n$ and let $x \cdot \omega$ denote the inner product $x_1\omega_1 \oplus ... \oplus x_n\omega_n$. Let $f(x)$ be a Boolean function in $n$ variables. Then the Walsh transform of $f$ is a real valued function over $\{0,1\}^n$ defined by 
    \begin{equation} \label{eq:4}
        W_f(\omega) = \underset{x\in\{0,1\}^n}{\sum}(-1)^{f(x)\oplus  x\cdot\omega}
    \end{equation}
\end{definition}

Let $x\cdot \omega$ be denoted by $l_\omega(x)$. Observe that the Walsh transform $W_f(\omega)$ computes the number of $x\in\{0,1\}^n$ for which $f(x)$ differs from $l_\omega(x)$, denoted by $\#(f \neq l_\omega)$, and subtracts it from the number of $x$ for which $f(x)$ equals $l_\omega(x)$, denoted by $\#(f = l_\omega)$. Also note that $\#(f = l_\omega) + \#(f \neq l_\omega) = 2^n$.

\begin{eqnarray} \label{eq:5}
   W_f(\omega) &= \#(f = l_\omega) - \#(f \neq l_\omega) \\
   &= 2^n - 2*\#(f \neq l_\omega) \\
   &= 2^n - 2*d(f, l_\omega)
\end{eqnarray}

This implies
\begin{equation} \label{eq:8}
   d(f, l_\omega) = 2^{n-1} - \frac{1}{2} W_f(\omega)
\end{equation}

For each $\omega \in \{0,1\}^n$,  $l_\omega$ denotes an affine function with the constant term set to 0. Observe that if we consider the set of affine functions with the constant term set to 1 instead, their corresponding Walsh transform values change in sign but remain the same in magnitude. (In this case, the Walsh spectrum would have to be calculated with  $f(x)\oplus  x\cdot\omega \oplus 1$ in the right hand side of Eq. \ref{eq:4}).

The set of values obtained by applying the Walsh transform of a function on all $\omega \in \{0,1\}^n$ is called its Walsh spectrum. Therefore, we can derive the nonlinearity of a Boolean function $f$ by calculating its Walsh spectrum.

\begin{equation}\label{eq:9}
    nl(f) = 2^{n-1} - \frac{1}{2} \underset{\omega \in \{0,1\}^n}{max}|W_f(\omega)|
\end{equation}

\subsection{Walsh Hadamard Transform}
\begin{definition}[Hadamard Matrix]
A Hadamard matrix of order k $(H_k)$ is any square matrix of size $k x k$ that has mutually orthogonal columns and elements from the set $\{+1, -1\}$.
\end{definition}

A particular series of Hadamard matrices called as the Walsh Hadamard matrices which can be constructed using the method as described below, help us calculate the Walsh spectrum of a given function and as a consequence, its nonlinearity. 

\begin{equation}
    H_2 = \begin{bmatrix}
            1 & 1\\
            1 & -1
        \end{bmatrix} 
\end{equation}

For $k\geq 2$ ($k$ must be a power of 2)

\begin{equation}
    H_{2k} = \begin{bmatrix}
                    H_{k} & H_{k}\\
                    H_{k} & -H_{k}
                \end{bmatrix}
\end{equation}

Given a function $f(x_1, ..., x_n)$, we can use the Hadamard Matrix to calculate its Walsh spectrum \cite{fino_algazi}. We first modify the truth table of $f$ by replacing $0$s and $1$s with $1$s and $-1$s respectively. If we denote the thus modified truth table as $[(-1)^f]_{N\times 1}$, the Walsh spectrum can be calculated as:

\begin{equation} \label{eq:12}
    [Walsh Spectrum]_{N\times 1} = H_{N} \times [(-1)^f]_{N\times 1}
\end{equation}

The Walsh spectrum of a function can be calculated asymptotically faster using the Fast Walsh Transform algorithm \cite{fino_algazi}. It runs in $O(NlogN)$ time compared to when using the Walsh Hadamard transform which takes $O(N^2)$ \hypertarget{section3}{time}.

\section{Learning Nonlinearity}
Nonlinearity of Boolean functions can be calculated with 100\% accuracy using different traditional methods. Consider $n$-variable Boolean functions whose truth tables are $N$($2^n$) bit strings. The Walsh spectrum can be calculated using the Walsh Hadamard matrix in $O(N^2)$ time or using the Fast Walsh Transform (FWT) in $O(Nlog(N))$ time. Finding nonlinearity from the Walsh spectrum can be done in linear time, therefore the best performing algorithm's time complexity is $O(Nlog(N))$. FWT does not need any additional space apart from that needed to store the function, therefore the algorithm requires $O(N)$ memory.

One way to reduce this time is to first calculate the nonlinearity of all the $2^N$ $n$-variable functions and store it in a database. This takes $O(2^NNlogN)$ time. Nonlinearity values have an upper bound which is in $O(N)$ and thus can be represented using $log(N)$ bits for each value. Therefore all the nonlinearity values can be stored using $O(2^Nlog(N))$ space. Once we have the completed calculating them, retrieving the nonlinearity of a particular function only takes $O(1)$ time. Since the function space grows super-exponentially in $n$, neither calculating nor storing nonlinearities of all the functions is feasible. 

Therefore, training a neural network on a fraction of the function space to learn a method to calculate nonlinearity that lies in between the above two methods, if not better than both, would be very useful. That is, one that takes lesser space than storing all the nonlinearities and is faster than the FWT algorithm during inference. A reasonably small error rate could be an acceptable compromise for some purposes. Before we set out to find such a method, the question of whether the property of nonlinearity is learnable at all is to be answered. We take two approaches to answer this question:

\subsubsection{Approach 1:} Learning the Walsh spectrum --- a simple linear transformation is all that is needed to get from Boolean functions to their Walsh spectra. Therefore, this problem can be formulated as learning the correct transformation matrix. 

This approach has the following limitation --- by only calculating the Walsh spectrum, we will still be one step away from calculating nonlinearity. Nonlinearity can be easily calculated from the Walsh spectrum using a regular program, but learnability by neural networks cannot be claimed directly since it involves operations like abs\_max which are known to be challenging for neural networks. Additionally, we would be training the network with the Walsh spectra of functions, which supplies the network with a lot more information compared to the single nonlinearity value which we want it to predict. Therefore we try out the second approach.

\subsubsection{Approach 2:} Learning nonlinearity end to end --- by only giving the Boolean functions and corresponding nonlinearity values as inputs to the network. 

Descriptions of the networks used, details about the experiments and results about learnability are given in the following \hypertarget{section3_1}{sections}.

\subsection{Learning the Walsh Spectrum}

\begin{figure}[h!]
\begin{center}
    \includegraphics[width=0.6\textwidth]{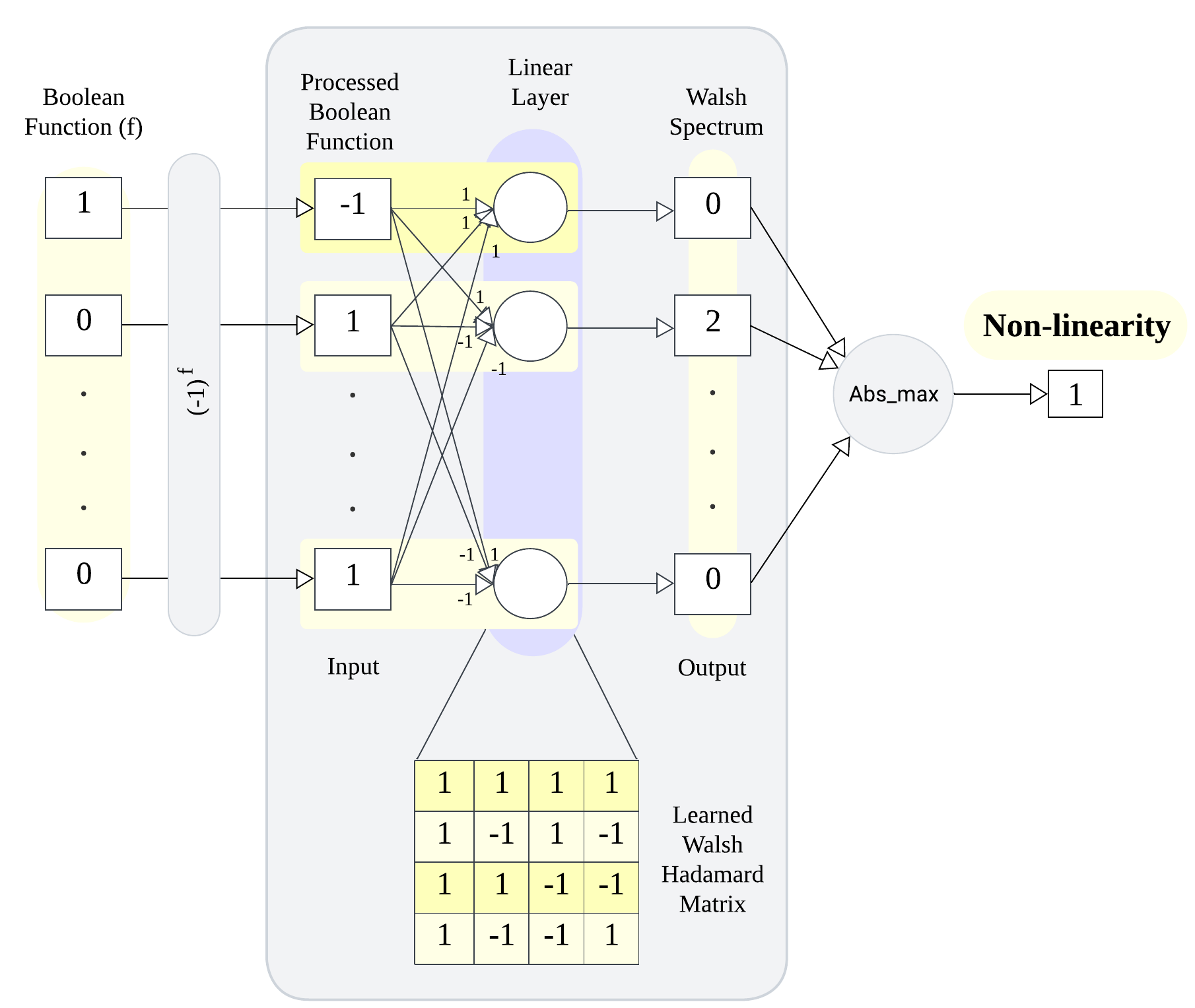}
\end{center}
\caption{Network used for learning the Walsh spectrum of $n$-variable Boolean functions. Processed Boolean functions (after replacing 0/1 in their truth tables with 1/-1) and corresponding Walsh spectra are used as examples to train the network. The weight matrix after convergence resembles the Walsh Hadamard matrix.} \label{fig1}
\end{figure}

It can be observed from Eqs. \ref{eq:12} and \ref{eq:9} that calculating the nonlinearity of Boolean functions is a combination of an affine transformation and an absolute max function. Therefore we first tried to learn the Walsh transform, which can be derived from a simple linear transformation from the processed Boolean function (see Fig. \ref{fig1}).

For a general $k$ layer feed-forward neural network, the output is given by
\begin{equation}\label{eq:13}
    \small f(x) = \sigma^{(k-1)}(b^{(k-1)} + W^{(k-1)}\sigma^{(k-2)}( ... \sigma^{(1)} (b^{(1)} + W^{(1)} x)))
\end{equation}
where for the $i$th layer (after the input layer), $\sigma^{(i)}$ is the activation function, $W^{(i)}$ is the matrix notation of its weights and $b^{(i)}$ is the vector notation of its bias. A network with a single layer without any activation functions, can however only act as a simple affine transformation:
\begin{equation}\label{eq:14}
    \small f(x) = b^{(1)} + W^{(1)} x    
\end{equation}

Since this is exactly what we need for modeling the Walsh spectra, we chose to train a single layered neural network with $N$ neurons (see Fig. \ref{fig1}) to learn the Walsh spectrum of $n$-variable Boolean functions. Comparing Eqs. \ref{eq:12} and \ref{eq:14} indicates that the weight matrix of such a network after training should resemble the Walsh Hadamard matrix of order $N$ and the bias vector should converge to the zero vector. This effectively converts the problem into the network learning the correct transformation matrix from the given input and output pairs - the solution being the Walsh Hadamard matrix of order $N$. Using a small number of $n$-variable functions as inputs and Walsh spectra as outputs, we were able to train the network to predict the Walsh spectrum perfectly (when rounded). The SGD optimizer was used for this purpose. Additionally, the weight matrix converged exactly to the Walsh Hadamard matrix. We tried this for functions in $n$=2 to 10 variables, and the same result was observed in all the cases. For example, for $n$=2 ($N$=4), the weight matrix converged to:

\begin{equation}weightMatrix = \begin{bmatrix}
1.001 & 0.995 & 1.000 & 1.006\\
1.001 & -1.007 & 0.999 & -1.001\\
1.000 & 1.000 & -1.004 & -1.005\\
0.995 & -0.995 & -0.997 & 1.003\\

\end{bmatrix},
H_4 = \begin{bmatrix}
1 & 1 & 1 & 1\\
1 & -1 & 1 & -1\\
1 & 1 & -1 & -1\\
1 & -1 & -1 & 1\\
\end{bmatrix}
\end{equation}

It was observed that for the case of $n$ variables, it was enough to train the network on just $N$ Boolean functions, under the condition that their vector forms are linearly independent. This observation nudged us to look at the problem from the angle of solving a set of linear equations.

Consider a Boolean function $f_0$ and its Walsh spectrum $W_0$ whose values are known. Assume that $H_N$ is unknown.

\begin{equation}\begin{bmatrix}
h_{11} & ... & h_{1N} \\
...& & ... \\ 
h_{N1} & ... & h_{NN} \\
\end{bmatrix} \times \begin{bmatrix}
    b_1\\...\\b_N
\end{bmatrix} = \begin{bmatrix}
    ws_1\\...\\ws_N
\end{bmatrix}\end{equation}

Here, we have $N^2$ unknowns of the form $h_{ij}$, where $i, j\in\{1, ..., N\}$. The vector equation $H_N \times f_0 = W_0$ for each function $f_0$ gives us $N$ regular equations of the form \begin{equation}b_1h_{i1} + ... + b_jh_{ij} + ... + b_Nh_{iN} = ws_i, i\in\{1, ..., N\}\end{equation} 
Therefore, if we use $N$ Boolean functions, which when expressed as vectors are linearly independent, we will get the $N^2$ equations that we need for a unique solution to the $N^2$ unknowns that comprise of the Hadamard matrix. This implies two points.

\begin{figure}
\begin{center}
    \includegraphics[width=0.4\textwidth]{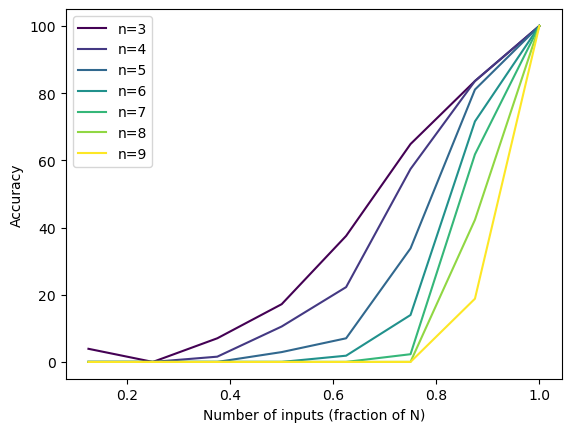}
\end{center}
\caption{Accuracy of the model drops significantly when less than $N$ linearly independent functions are given as inputs to the linear network. The drop gets sharper as $n$ increases.} \label{fig3}
\end{figure}

\begin{itemize}
    \item This task of solving a set of linear equations (for which we know that there is a solution in the form of a Hadamard matrix) is being modeled by us as a linear regression problem by choosing to use a linear network to solve it. Also, linear regression is known to be a convex optimization problem. This means that finding the solution is a matter of performing a simple gradient descent on the convex surface defined by the loss function. This indicates that our findings can be extended to Hadamard matrices of higher order and thus Boolean functions of higher arity.

    \item If we use less than $N$ Boolean functions, then multiple solutions exist for the transformation matrix. That means, to be sure that we arrive at the Hadamard matrix $H_N$ as the solution, we need to train the model on at least $N$ linearly independent Boolean functions. This result was experimentally verified (see Fig. \ref{fig3}) by training with less than $N$ functions.
\end{itemize}

It was observed that $N$ linearly independent inputs were enough to train the network, but using a larger number of inputs helps the network to converge faster (Up to 100x speedup was observed for larger values of $n$).

This approach has the advantages of providing us with an interpretable network, giving us some guarantees about extending the positive results to higher number of variables etc., but this only gets us till the Walsh spectrum. Therefore, to address the limitations discussed here and in the beginning of section 3, we move on to attempting to learn nonlinearity in an end to end \hypertarget{section3_2}{fashion}.


\subsection{Learning Nonlinearity End-to-end}

\begin{figure}[h!]
\begin{center}
    \fbox{\includegraphics[width=0.45\textwidth]{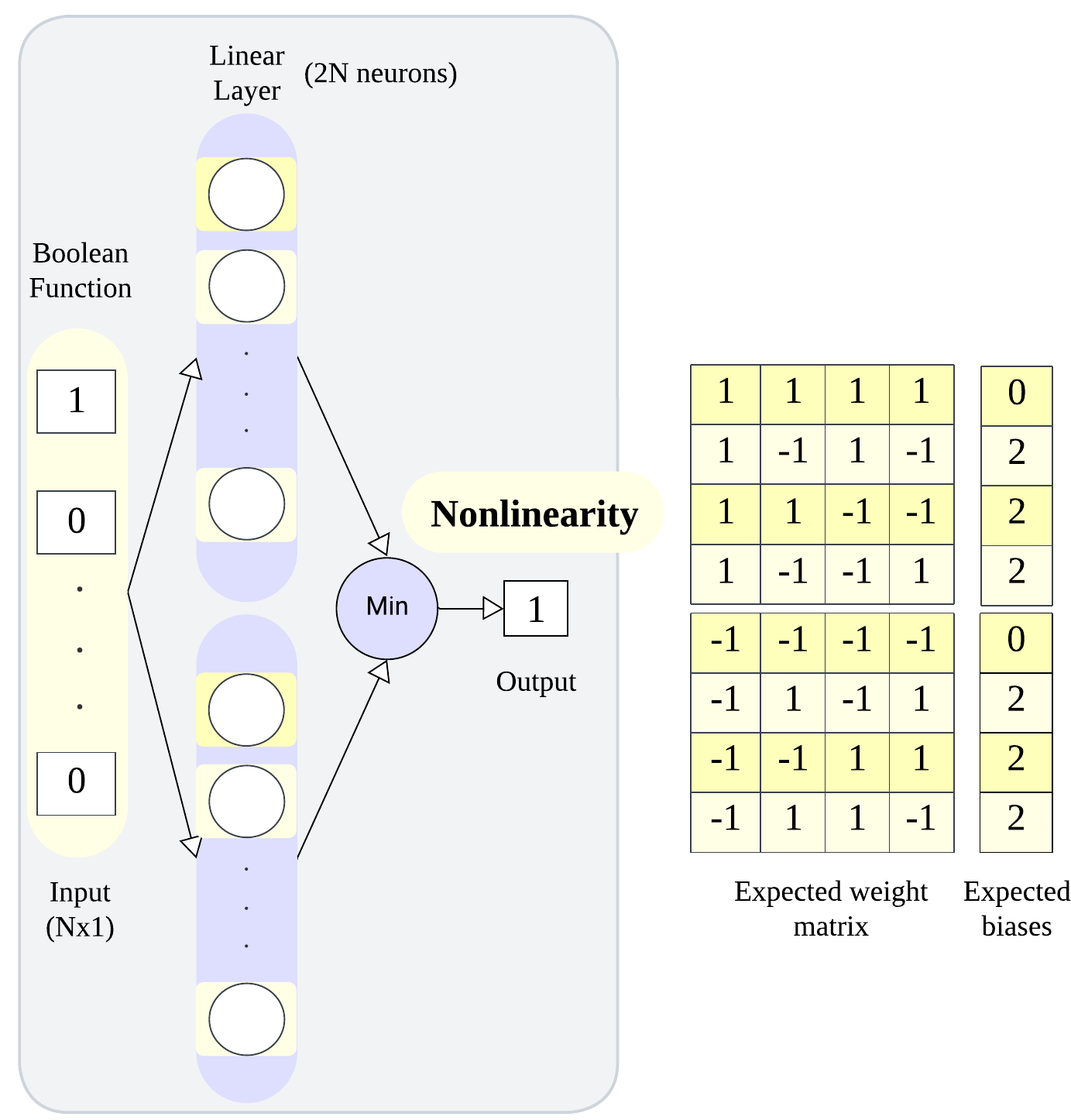}}
    \fbox{\includegraphics[width=0.453\textwidth]{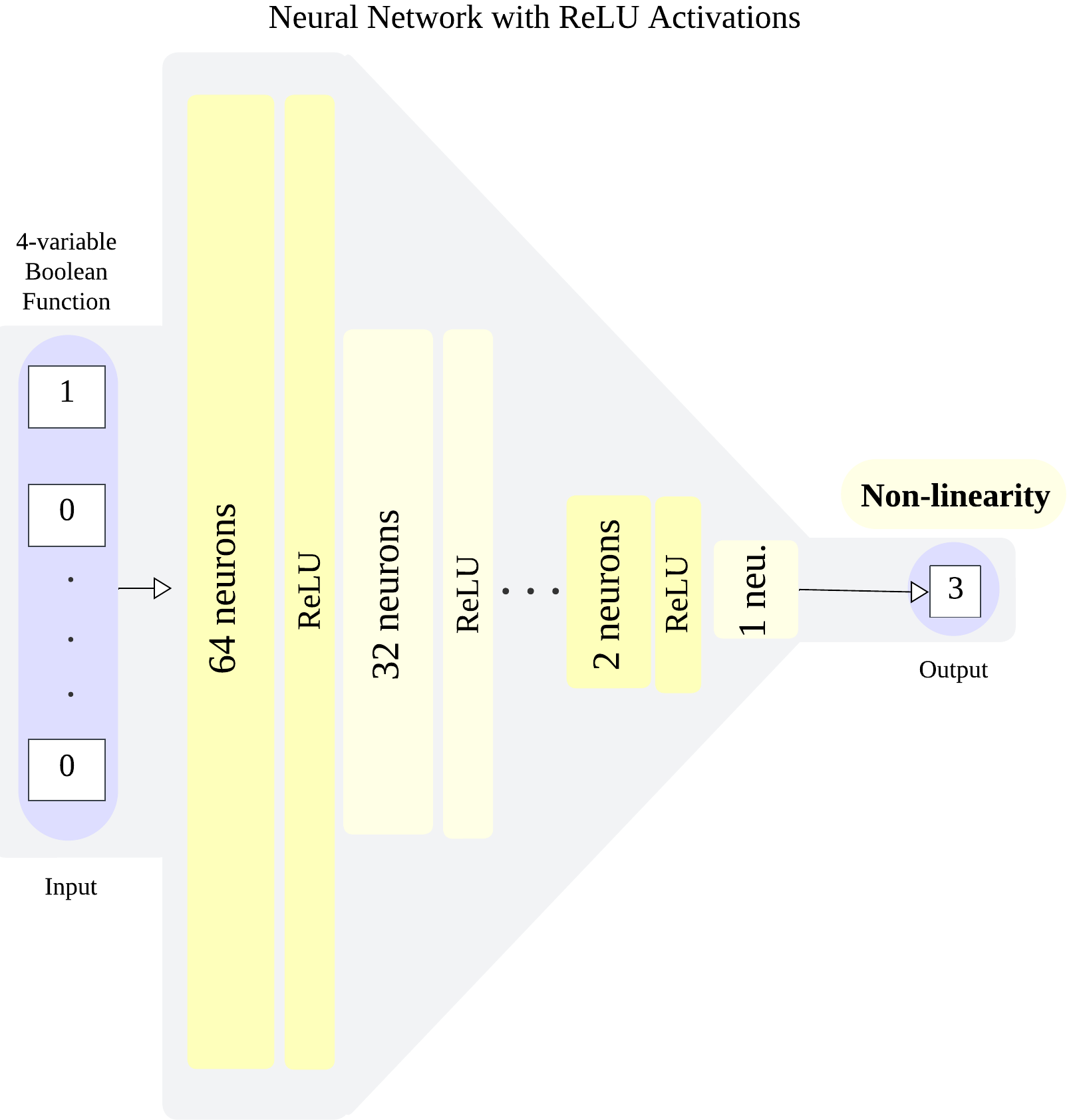}}
\end{center}
\caption{Networks used for learning the nonlinearity of Boolean functions in an end to end fashion. Only Boolean functions and corresponding nonlinearity values are used as examples to train the network. On the left is a network that can model nonlinearity calculation, but does not respond well to training. As an example, expected weights for 2-variable functions are given. On the right is the deep encoder style network using which we could learn to predict nonlinearity. The design used for 4-variable functions is shown in the figure. } \label{fig2}
\end{figure}

Note that the nonlinearity of a given $n$-variable function can be calculated using the distances of the given function from all the $2N$ affine functions (which can be calculated using an affine transformation according to Eq. \ref{eq:8}) and finding their minimum. Additionally, the processing step of going from $f$ to $(-1)^f$ can be replaced by $(1-2f)$ since we are only dealing with the values 0/1. This means that the whole process of calculating nonlinearity can be simplified to a suitable affine transformation followed by finding its minimum. 
Therefore, we first tried to use a neural network similar in structure to the linear one used for learning the Walsh spectrum, but only the nonlinearity values are given to the network. Twice the number of neurons were used, so that the network could calculate the distances from all the $2N$ affine functions, as opposed to the Walsh spectra corresponding to the $N$ linear functions. See the network in the left half of Fig. \ref{fig2} for the expected values of the parameters. The min operation was implemented using a Convolutional Neural Network style 1D max\_pool layer.

This network with the given values can model nonlinearity calculation, but we were unable to train the network to converge to these values (or any set of values equivalent to this i.e. with the same rows of the weight matrix but in any of the ($2N$)! permutations). In order to learn to predict nonlinearity, when given some examples, the network should move all its parameters towards the values of the rows of the Walsh Hadamard matrix which correspond to each of the affine functions. All this information needs to be gathered from just the Boolean functions and nonlinearity values (as opposed to the more informative Walsh spectra). We did not expect that the loss surface for the network and problem would be in such a way that gradient descent would lead stably to any of the ($2N$)! global minima, and that is what we observed as well.

But neural networks are known to improve in performance as the number of neurons and depth of the network increases. Therefore, we tried out encoder style fully connected networks (see the right half of Fig. \ref{fig2}) of depth and width much larger than that in the above network. We chose a base width (value decided by trial and error) for the first layer. The width of each subsequent layer decreases by half until we reach the layer with only one neuron that gives us the nonlinearity. We used ReLU activation functions in between the layers. With this design, were able to train networks to learn to predict nonlinearity for functions in 4 and 5 variables (we failed to do so for functions in 3 variables, possibly because of the small size (256 functions) of the function space). Refer to Table \ref{tab:network_structure} for the exact structure of the networks used.

\begin{table}[h]
\centering
\caption{Structure of networks used for learning nonlinearity end to end. Number of neurons used in each layer and total number of parameters is given.}
\label{tab:network_structure}
\resizebox{\columnwidth}{!}{%
\begin{tabular}{ccccc}
\hline
$n$ & input layer & hidden layers                      & output layer & total parameters \\
\hline
4 & 16          & 64, 32, 16, 8, 4, 2                & 1            & 3881             \\
\hline
5 & 32          & 512, 256, 128, 64, 32, 16, 8, 4, 2 & 1            & 192196            \\ 
\hline
\end{tabular}
}
\end{table}

\begin{figure}
\begin{center}
    \fbox{\includegraphics[width=0.473\textwidth]{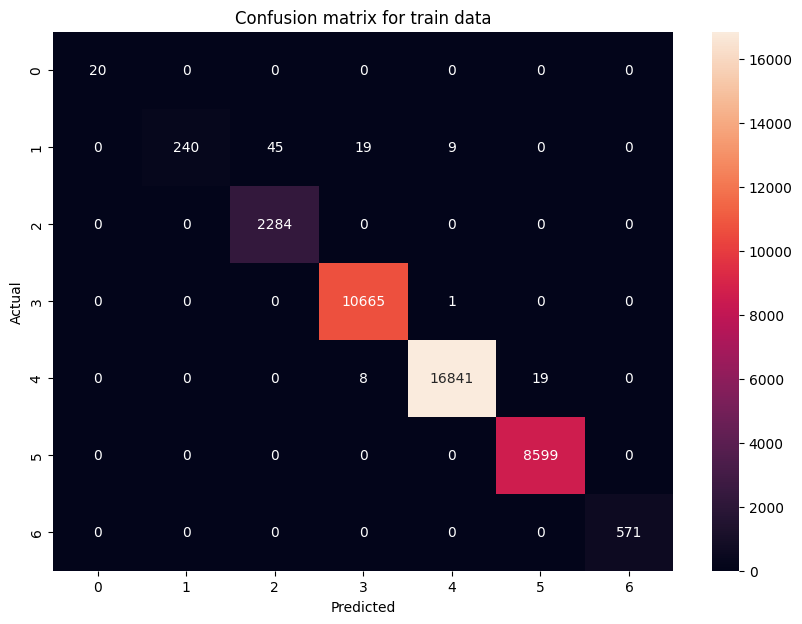}}
    \fbox{\includegraphics[width=0.47\textwidth]{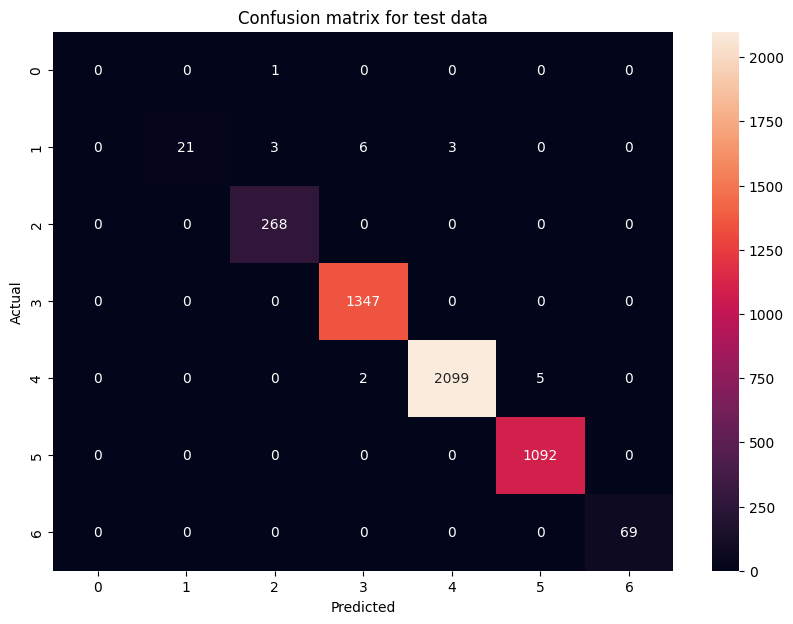}}
\end{center}
\caption{Confusion matrices of the predictions made by the model trained on 4-variable Boolean functions in an end to end fashion. Predictions for train and test data are given.} \label{fig4}
\end{figure}

For example, for functions in 4 variables, we chose a network with 7 layers (excluding the input layer) with the following number of neurons: 64, 32, 16, 8, 4, 2, 1. We were able to train the network (4k parameters) with around half the function space (30k examples) and were able to achieve 99.7\% train and 99.5\% test accuracy. See Fig. \ref{fig4} for the confusion matrices on train and test data. Adding another layer with 128 neurons at the beginning did not give any increase in the accuracy, but removing layers resulted in a massive drop to around 40\%. We were able to extend the results to functions in 5 variables and reach 98.2\% train and 96.2\% test accuracy, but we needed to scale up to a much larger network (200k parameters, 10 layers starting from 512 neurons in the first layer) and a much higher number of examples (200k). We were unable to extend this further to 6 or higher variable functions, even with the maximum capacity of 20x more parameters, 2x layers and 5x examples that our system allowed us to go till. Size and complexity of neural networks is usually measured by the number of parameters and there is no way to theoretically predict the size of the model and training data needed to train a network well, but the above experiments indicate that the requirements grow very quickly for this problem of learning nonlinearity. The fact that the function space grows exponentially in $N$ provides some support for this observation.

For $n$=4 and 5, given that the width of the first hidden layer of the network was required to be more than $N$, Eq. \ref{eq:13} implies that calculating predictions using the trained network involves vector-matrix multiplications with matrices that are larger than $(N\times N)$. Therefore, the network performs worse than the simple Walsh Transform algorithm (which by itself is slower than FWT). The memory required is also clearly more than $N$ units. Therefore, we have failed to come up with networks that are better that traditional algorithms in either memory or \hypertarget{section4}{speed}.

\section{Conclusions}

Learning the nonlinearity property in an end to end manner is challenging. The task is feasible for Boolean functions of 4 and 5 variables, but the resources required for training grow rapidly after that and in this paper, we did not find a feasible solution for functions in higher variables. Additionally, the performance of the model during inference is not as good as that of traditional algorithms. 

In the future, one can explore the possibility of using neural networks to perform nonlinearity testing where the task is to predict whether a given Boolean function is linear or not by seeing as few entries of its truth table as possible.

\bibliographystyle{spmpsci}
\bibliography{learnability}

\end{document}